\documentclass[10pt,conference]{IEEEtran}

\usepackage{booktabs}
\usepackage{amsmath}
\usepackage{graphicx}
\usepackage{cite}
\usepackage{threeparttable}
\usepackage[hidelinks]{hyperref}
\hypersetup{pdfauthor={}, pdftitle={}, pdfsubject={}, pdfkeywords={}}

\newif\ifanon
\anonfalse

\begin{document}

\title{Evaluating Tiny Recursive Models\\Across Training for Code Generation}

\ifanon
  \author{\IEEEauthorblockN{Anonymous Author(s)}
          \IEEEauthorblockA{Paper submitted for double-anonymous review}}
\else
  \author{%
    \IEEEauthorblockN{Anjani Sirivella}
    \IEEEauthorblockA{\textit{Dept.\ of Computer Science} \\
                      \textit{Toronto Metropolitan University} \\
                      Toronto, ON, Canada \\
                      asirivella@torontomu.ca}
    \and
    \IEEEauthorblockN{Aanisha Newaz}
    \IEEEauthorblockA{\textit{Dept.\ of Computer Science} \\
                      \textit{Toronto Metropolitan University} \\
                      Toronto, ON, Canada \\
                      aanisha.newaz@torontomu.ca}
    \and
    \IEEEauthorblockN{Glaucia Melo}
    \IEEEauthorblockA{\textit{Dept.\ of Computer Science} \\
                      \textit{Toronto Metropolitan University} \\
                      Toronto, ON, Canada \\
                      glaucia@torontomu.ca}
  }
\fi

\maketitle

\begin{abstract}
Code generation increasingly relies on large transformer models, whose capability advances with scale. Yet such a scale is costly, creating demand for small models,
especially where data is limited. Recursive models address this by reusing a single
block to add depth rather than stacking independent layers. Such models are typically
evaluated by teacher-forced fit (next-token loss on ground-truth prefixes) or task
accuracy, at a single checkpoint, whereas code is produced by
free-running generation, where the model extends its own output. Whether a teacher-forced advantage survives free-running
generation, and whether it holds across training, remains open. To study both, we compare a $\sim\!28$M-parameter
autoregressive Tiny Recursive Model (TRM-AR) on natural-language-to-Python code generation against parameter-matched and depth-matched controls, tracking fit and generation across 40 epochs and three seeds. The fit ranking between the recursive model and the depth-matched control reverses twice. Selecting each checkpoint
by validation loss and examining the trajectory yields a consistent comparison. At equal parameters, TRM-AR fits, generates, and generalizes better than the parameter-matched control while recovering approximately 45\% of the validation-loss gap and 57\% of the generation-quality gap between the two controls, at roughly $175\times$ the per-step cost of the parameter-matched control. However, at equal effective depth, the larger transformer fits and generates better at its validation optimum, suggesting TRM-AR's advantage lies in resistance to overfitting, not greater capability. These findings suggest that recursive code generation models should be evaluated jointly on fit and generation across the training trajectory rather than at a single checkpoint.
\end{abstract}

\section{Introduction}
\label{sec:intro}

Code generation from natural language has become a central task in software
engineering~\cite{codegensurvey}. The systems that dominate it are transformer-based
language models~\cite{vaswani}, whose capability has advanced largely by scaling
parameters and data~\cite{chinchilla}. State-of-the-art code models now reach
billions of parameters, trained on vast code corpora~\cite{codegen,codet5plus,alphacode,starcoder}.
Models at this scale are costly to train and to serve~\cite{schwartz2020green},
which creates practical demand for small models that generate code well. That
demand is most acute where training data is limited, since additional parameters
are hardest to feed in exactly that regime~\cite{muennighoff2023scaling}. Whether small models can generate code well without scaling parameters remains open.

Recursive models address this directly: they obtain computational depth by reusing a
single small block of weights rather than by stacking independent
layers~\cite{ut,albert}. The Tiny Recursive Model (TRM)~\cite{trm}, a minimal
simplification of the Hierarchical Reasoning Model~\cite{hrm}, is a striking
instance: with only 7M parameters, it reaches roughly 45\% test accuracy on
ARC-AGI-1, reportedly surpassing far larger language models such as DeepSeek R1, o3-mini,
and Gemini 2.5 Pro~\cite{trm}. Yet the literature evaluates this family almost
entirely through \emph{teacher-forced fit} (next-token prediction from
ground-truth prefixes)~\cite{teacherforcing}, or task accuracy~\cite{saunshi}; for TRM and its predecessor specifically, that accuracy comes from puzzle tasks whose answers
are scored by exact match, reported at a single converged checkpoint~\cite{trm}.
Generating code, however, requires \emph{free-running} generation, in which the
model conditions on its own earlier output rather than on a ground-truth
prefix~\cite{professorforcing}.

We argue that the comparison between recursive and standard models is time-indexed: which model performs better depends on what is measured and when in training it is read. A teacher-forced advantage need not
reach free-running generation, because the two regimes are known to diverge~\cite{nexttokenpitfalls} and because early errors can compound over a long
generated sequence~\cite{exposurebias}. Model behaviour also changes across
training, so the point at which a comparison is read can change its
conclusion~\cite{pythia}. Teacher-forced fit and free-running generation must
therefore be tracked jointly across the whole training trajectory. \looseness=-1

In this paper, we instantiate this evaluation as a controlled, multi-epoch
\emph{crossover-curve} study of natural-language-to-Python code generation at
$\sim\!28$M parameters on the tiny-codes corpus~\cite{tinycodes}. A recursive
model differs from a plain transformer on two axes at once: parameter count and
effective depth. We therefore compare an autoregressive adaptation of TRM
(TRM-AR) against two controls that each hold one axis fixed: a transformer
matched to its parameter budget (\emph{iso-parameter}) and a transformer matched
to its effective depth (\emph{iso-depth}), the latter necessarily larger. We
evaluate all three across 40 epochs and three seeds, measuring teacher-forced
fit separately from free-running generation.

Across training, the fit ranking between TRM-AR and the iso-depth control is not stable across checkpoints: it reverses twice within the budget. 
The latter reversal is an overfitting effect, arising as the iso-depth control's validation fit worsens while the recursive model holds steady. We select each model's checkpoint on validation loss and read the whole trajectory, which gives two stable comparisons a single checkpoint cannot: one at equal parameters and one at equal effective depth.

Single-checkpoint comparisons of recursive models can therefore be artifacts of
the training time at which they are read. We accordingly recommend that
evaluations of such models report teacher-forced fit
and free-running generation jointly across the training
budget, echoing broader calls to strengthen empirical
rigor in machine learning~\cite{herrmann2024rethinking}.

We ask the following research questions (RQs). The first concerns how the comparison evolves across training; the remaining three examine parameter efficiency, compute cost, and generation quality.
\begin{itemize}
  \item \textbf{RQ1:} How does the recursive depth's advantage over a depth-matched, larger
        transformer change across the training budget?
  \item \textbf{RQ2:} How does recursive depth affect parameter efficiency for
        code generation at equal parameters?
  \item \textbf{RQ3:} How does the per-step compute cost of recursion compare to
        the fit gain it yields at equal parameters?
  \item \textbf{RQ4:} How does recursive depth affect free-running generation
        quality at equal parameters?
\end{itemize}

This paper makes the following contributions:
\begin{itemize}
  \item A multi-epoch \emph{crossover-curve} evaluation that tracks
        teacher-forced fit against free-running generation across training and
        three seeds, with parameter- and depth-matched controls; we show that a single checkpoint can produce a
        misleading ranking, as several epochs here do, which reading the full trajectory corrects.
  \item Evidence that, at equal parameters, the TRM-AR configuration (recursive
        depth with deep supervision) improves fit, generation as scored by an
        LLM judge~\cite{llmjudge}, \emph{and} generalization from early training
        onward, extending the established parameter-efficiency claim for looped
        models to free-running code generation (Section~\ref{sec:results}).
  \item A boundary result: at equal effective depth, a larger plain transformer both
        fits and generates better at its optimum, so what recursion retains
        against it is resistance to overfitting (consistent with implicit
        regularization), not superior capability (Section~\ref{sec:results}).
  \item A measurement of the per-step compute cost of recursion, showing that it
        is parameter-efficient but not compute-efficient
        (Section~\ref{subsec:compute}).
  \item A replication package with code, configuration, and the
        aggregated crossover-curve results is publicly available~\cite{artifact}.
\end{itemize}

The remainder of this paper is organized as follows. Section~\ref{sec:relwork}
reviews related work. Section~\ref{sec:method} describes the experimental
methodology. Section~\ref{sec:results} presents the results and
Section~\ref{sec:discussion} interprets them.
Section~\ref{sec:scope} states the limitations of this study and
Section~\ref{sec:conclusion} concludes the paper.

\section{Related Work}
\label{sec:relwork}

This work draws on two bodies of prior work: recursive, weight-shared architectures, which supply the models we compare, and evaluation research on metric and checkpoint sensitivity, which shapes how we compare them.

\subsection{Recursive and Weight-Shared Transformers}
Obtaining depth by reusing a shared block, rather than by stacking independent
layers, is a long-standing idea. The Universal Transformer applies one shared block
recurrently across depth and is, under certain assumptions,
Turing-complete~\cite{ut}. ALBERT popularized cross-layer parameter sharing to
reduce model size~\cite{albert}. Looped-block comparisons made the substitution
quantitative: a $k$-layer block looped $L$ times nearly matches a $kL$-layer
non-looped model on reasoning tasks, at the parameters of the $k$-layer
block~\cite{saunshi}. Relaxed Recursive Transformers convert pretrained models to
weight-shared form and recover most of the original quality once the sharing
constraint is relaxed~\cite{relaxedrecursive}. Recurrent-depth models unroll a
shared block at inference, trading additional computation for improved reasoning
without emitting intermediate tokens~\cite{geiping}. The TRM, which we adapt, differs from the works above by adding deep supervision over iterative latent refinement rather than reusing weights alone. It is a deliberately minimal instance, simplifying the Hierarchical Reasoning Model (HRM)~\cite{hrm} to one repeatedly
applied two-layer block~\cite{trm}.

These works report two kinds of evidence, both at converged checkpoints.
Where the models are small, as with TRM and HRM, the results are teacher-forced fit
or accuracy on one-shot puzzles; generation has been examined at a billion-parameter
scale~\cite{ouro} and, in concurrent work, on synthetic character-level algorithmic
tasks~\cite{artrm}. 
Neither shows whether the fit advantage that motivates these architectures survives free-running generation, where the model extends its own earlier output rather than a ground-truth prefix, for natural-language-to-code, or across training. 
That question is most pressing at the small scale, where parameter efficiency is the primary motivation. \looseness=-1

\subsection{Depth through Recursion versus Depth through Parameters}
Across these works runs one substitution claim, that a shallow block, looped, can
stand in for stacked layers, obtaining depth at a fraction of the parameter
count~\cite{saunshi,relaxedrecursive,geiping}. Testing this claim requires matched
controls, because a recursive model differs from a plain transformer on two axes at
once, parameter count and effective depth. The closest prior work already uses two matched controls: Saunshi et al.~\cite{saunshi} compare a looped block against an iso-parameter shallow
model and against an iso-FLOP model of equal depth but with many more
parameters. Concurrent work also adapts TRM to autoregressive
decoding, finding no reliable gains under compute-matched comparisons on synthetic character-level algorithmic tasks~\cite{artrm}. 

Our adaptation differs in three ways. First, it retains TRM's
deep-supervision training procedure. Second, whereas prior comparisons
report fit or accuracy at a single converged checkpoint, implicitly
assuming the ordering is stable across training, we track both matched
comparisons across the whole trajectory and evaluate free-running
generation alongside teacher-forced fit, on natural-language-to-code.
Third, following Saunshi et al.'s two-control logic, we match parameters
and \emph{effective depth} rather than compute, and treat per-step cost as
a measured quantity (Section~\ref{subsec:compute}) rather than a matched one. We do not label this control ``iso-FLOP'': although TRM-AR and the
depth-matched control perform comparable FLOPs in a single forward
pass, with both applying twenty transformer layers, TRM-AR's measured per-step cost is higher, because deep supervision applies the recursion over three
supervision steps per optimizer step (Section~\ref{subsec:compute}). The larger depth-matched control is expected
to fit better at its validation optimum, since parameter count, not depth alone,
governs how much a model can store~\cite{knowledgecapacity}. Our contribution is to reshape that ranking across training and under generation.

\subsection{Metric and Checkpoint Sensitivity in Evaluation}
Teacher-forced next-token prediction can diverge from free-running generation, so low
held-out loss does not by itself imply good generated output~\cite{nexttokenpitfalls}. This train--inference mismatch was recognized early
in neural sequence modelling and motivated corrections such as scheduled sampling
and sequence-level training~\cite{scheduledsampling,seqleveltraining}. For code in
particular, the standard target is execution-based functional correctness such as
pass@$k$~\cite{codex}; where no unit tests are available, reference-free
LLM-as-judge scoring has become a common proxy for generation
quality~\cite{llmjudge}. Checkpoint suites such as Pythia~\cite{pythia} show that model behaviour
can change qualitatively over the course of training, so a single checkpoint can
misrepresent a model. This risk is amplified in our setting because
small models are typically trained for multiple passes over a limited corpus. In
this repeated-data regime, repeated tokens and excess parameters both lose
value~\cite{muennighoff2023scaling}, and models of different capacity need not
overfit on the same schedule. 

To our knowledge, no prior work charts how a
recursive, weight-shared model and its parameter- and depth-matched controls
compare on \emph{both} teacher-forced fit and free-running generation \emph{as
training proceeds}, across multiple seeds. This paper charts that trajectory: where the three models cross, plateau, or diverge and, consequently, where a ranking read
from an arbitrarily chosen checkpoint could be an artifact of the epoch rather than a property of the architectures.

\section{Methodology}
\label{sec:method}
We describe the experimental setup used to compare recursive
and non-recursive models on natural-language-to-Python code generation. We evaluate three models, which we call the arms of the study: TRM-AR, a
recursive model, and two non-recursive transformer controls that each match it
on one axis, parameter count (iso-parameter, vanilla-2L) or effective depth
(iso-depth, vanilla-20L). 
\iffalse
All arms share one corpus (Section~\ref{subsec:dataset}), the same transformer block
(Section~\ref{subsec:models}), and one optimization protocol
(Section~\ref{subsec:training}), so that measured differences reflect how depth is
realized. This does not isolate depth from the deep supervision the recursive arm
uses (Section~\ref{sec:scope}). We
evaluate every arm along the training trajectory rather than at a single
checkpoint, separating teacher-forced fit from free-running generation
(Section~\ref{subsec:eval}), and we treat recursion's per-step compute as a
measured quantity (Section~\ref{subsec:compute}).
\fi

\subsection{Dataset}
\label{subsec:dataset}

All three arms are trained and evaluated on a single shared corpus, holding the data fixed across arms. The corpus
is derived from \textsc{tiny-codes}~\cite{tinycodes}, a synthetic, LLM-generated
collection of short instruction--code snippets created in the ``textbook-quality''
style~\cite{phi1}. The full release contains roughly $1.6$M snippets
spanning a dozen programming and query languages. Each record pairs a
natural-language instruction (a \texttt{prompt}) with a \texttt{response}
containing the corresponding code. \looseness=-1

\paragraph{Subset and preprocessing}
We first restrict the corpus to its Python records, yielding $129{,}063$
prompt--response pairs. We tokenize with the GPT-2 byte-pair
encoder (vocabulary size $50{,}257$)~\cite{gpt2} and discard any record whose
tokenized \texttt{prompt}--\texttt{response} length exceeds the $512$-token
context window ($81{,}414$ records removed). An exact-duplicate guard over
(\texttt{prompt}, \texttt{response}) pairs is then applied and removes none. The pipeline therefore
retains $47{,}649$ records ($36.9\%$ of the Python subset), which constitute the
working corpus.

\paragraph{Splits}
We partition the filtered corpus into training, validation, and test sets in an
$80/10/10$ ratio under a fixed seed ($42$), yielding the counts in
Table~\ref{tab:splits}. This gives held-out validation and test sets of
$4{,}765$ examples each, large enough for stable checkpoint selection and
evaluation.

\begin{table}[t]
\centering
\renewcommand{\arraystretch}{1.3}
\begin{threeparttable}
\caption{Corpus Splits}
\label{tab:splits}
\setlength{\tabcolsep}{12pt} 
\begin{tabular}{lr}
\toprule
Split & Records \\
\midrule
Train & 38{,}119 \\
Validation & 4{,}765 \\
Test & 4{,}765 \\
\midrule
Total & 47{,}649 \\
\bottomrule
\end{tabular}
\begin{tablenotes}[flush]
\footnotesize
\item Python subset, records $\leq\!512$ tokens, split $80/10/10$ at seed~$42$.
\end{tablenotes}
\end{threeparttable}
\end{table}

\paragraph{Leakage control}
Held-out evaluation is informative only if the test split is free of training
contamination. Exact cross-split leakage of (\texttt{prompt}, \texttt{response}) pairs is
precluded by construction: deduplication is applied before the split, so
the test split is drawn from a duplicate-free pool. In addition,
the models are trained only on the training splits. The validation split is reserved for checkpoint selection and the test split for final evaluation. Empirically, prompt-level overlap between the test split and
the other splits is zero: none of the $4{,}765$ test prompts appears in the
training or validation splits under either exact or whitespace-normalized
matching, and all $4{,}765$ test prompts are mutually distinct. However, we do not
measure near-duplicate (fuzzy) or semantic overlap. Exact matching is a
deliberately conservative floor rather than a guarantee against paraphrastic
similarity, and this residual is stated as a limitation (Section~\ref{sec:scope}).

\paragraph{Scope of the corpus}
We use \textsc{tiny-codes} as a controlled testbed for an architecture comparison
on a shared instruction-to-code corpus, not as a benchmark of absolute
code-generation quality. Its synthetic, templated style is clean and learnable at
the $\sim\!28$M-parameter scale under study; the corresponding external-validity
limitation relative to human-authored code is discussed in
Section~\ref{sec:scope}.

\subsection{Models}
\label{subsec:models}

All three arms are built from an \emph{identical} transformer block and differ
only in how computational depth is realized: by recursively reusing a shallow
block, or by stacking independent layers. Fixing the block this way removes
incidental architectural differences as a confound. Notably, it does not by
itself separate recursion from the deep supervision that the recursive arm also
uses (Section~\ref{sec:scope}).

\paragraph{Shared backbone}
The basic unit is a decoder-only transformer layer with model
dimension $256$ and $8$ attention heads (head dimension $32$). Each layer uses a
SwiGLU feed-forward sublayer~\cite{swiglu} of hidden width $1024$ ($4\times$ the
model dimension), post-norm RMSNorm~\cite{rmsnorm} (with $\epsilon=10^{-6}$),
rotary position embeddings~\cite{rope} (no absolute position
embeddings), and causal self-attention. Inputs use a context length of
$512$; all weights are initialized from $\mathcal{N}(0,0.02)$. Two such layers
constitute the two-layer block used throughout.

\paragraph{Arms and matched controls}
To separate parameter count from effective depth, we compare TRM-AR against two matched controls. Table~\ref{tab:arms} lists each
arm's architecture and parameter counts:
\begin{itemize}
  \item \textbf{TRM-AR} realizes an effective depth of $20$ by recursively reusing
        the shared two-layer block (Eq.~\ref{eq:effdepth}), at essentially the
        parameter budget of a two-layer model.
  \item \textbf{vanilla-2L} (\emph{iso-parameter} control): two independent layers,
        matching TRM-AR's parameter budget to within $768$
        parameters (under $0.003\%$), which are TRM-AR's learnable initial states
        $y_0$ and $z_0$ and its (disabled) halting head, each a
        $256$-dimensional tensor. 
  \item \textbf{vanilla-20L} (\emph{iso-depth} control): twenty independent layers,
        matching TRM-AR's effective depth of $20$ at $1.68\times$ the total, or
        $\approx\!10\times$ the non-embedding, parameters~\cite{kaplan}.
\end{itemize}

\begin{table*}[t]
\centering
\renewcommand{\arraystretch}{1.3}
\setlength{\tabcolsep}{10pt}
\begin{threeparttable}
\caption{Architecture and Parameter Counts of the Three Arms}
\label{tab:arms}
\begin{tabular}{lccrr}
\toprule
Arm & Depth realization & Eff.\ depth & Parameters & Non-emb.\ params\tnote{a} \\
\midrule
TRM-AR      & recursive shared $2$-layer block & $20$ & 27{,}830{,}784 & 2{,}099{,}200 \\
vanilla-2L  & $2$ independent layers            & $2$  & 27{,}830{,}016 & 2{,}098{,}432 \\
vanilla-20L & $20$ independent layers           & $20$ & 46{,}713{,}600 & 20{,}982{,}016 \\
\bottomrule
\end{tabular}
\begin{tablenotes}[flush]
\footnotesize
\item Counts are exact.
\item[a] Trainable non-embedding parameters; embedding tables are identical across arms.
\end{tablenotes}
\end{threeparttable}
\end{table*}

\paragraph{Recursive computation}
TRM-AR adapts TRM~\cite{trm} to autoregressive generation. The
model maintains two states, a latent $z$ and an answer $y$, each initialized
from a learned parameter ($z_0$, $y_0$). One \emph{improvement cycle} performs
$n=4$ latent updates, each refining $z$ from an additive combination of the input
representation, the current answer, and the current latent, followed by a single
update that refines the answer $y$ from the latent. $T=2$ such cycles are applied
per step. 
We define \emph{effective depth} as the number of times the shared two-layer
block ($\ell=2$) is applied per forward pass:
\begin{equation}
\label{eq:effdepth}
D_{\text{eff}} \;=\; T\,(n+1)\,\ell \;=\; 2\times(4+1)\times 2 \;=\; 20,
\end{equation}
Training uses deep supervision over $n_{\text{sup}}=3$ steps: at each
step the model emits logits and incurs a next-token cross-entropy, and the
training loss is the mean over the three steps~\cite{trm}. Between supervision
steps the states $y$ and $z$ are detached. Within a step, the gradient uses a
one-step approximation: the first $T\!-\!1$ cycles run without gradient tracking
and only the final cycle is differentiated, which keeps the training cost of
recursion bounded.

\paragraph{Inactive and omitted features}
The halting head is present but inactive: adaptive computation is switched off
(its loss weight is $0$) so that every arm executes a fixed compute budget.
Because the recursive states are recomputed from scratch on each forward
pass, TRM-AR maintains no key/value cache during generation; the full recursion is
re-executed per generated token.

\subsection{Training}
\label{subsec:training}

All three arms are trained under an identical optimization protocol; only the
architecture (Section~\ref{subsec:models}) differs. This keeps the comparison a
function of architecture rather than of tuning, and makes checkpoints at matched
epochs directly comparable across arms. \looseness=-1

\paragraph{Optimizer}
We optimize with AdamW~\cite{adamw} ($\beta_1{=}0.9$, $\beta_2{=}0.95$, weight
decay $0.1$) at a peak learning rate of $1\times10^{-4}$, with gradients clipped
to a global norm of $1.0$. All runs use a batch size of $48$ and are trained in
single-precision (fp32).

\paragraph{Learning-rate schedule}
The learning rate follows a \emph{flat-after-warmup} schedule, with a linear warmup
over the first $2{,}000$ optimizer steps to the peak rate, followed by a constant
rate for the remainder of training. A decaying schedule (e.g.\ cosine) instead lowers the learning rate toward a small final value over a set number of training steps, which we avoid here for two reasons. First, a decay schedule would have to be
committed to a fixed horizon, and its intermediate checkpoints would fall at
systematically different points of the decay curve, confounding any comparison
of models \emph{across} training. Second, a decaying rate drives the loss change
toward zero and can mimic convergence. Holding the rate constant after warmup
instead places every snapshot in the same optimization regime: the ``stable''
phase of the increasingly standard warmup--stable--decay family~\cite{wsd}. We
omit the terminal decay phase and maintain an exponential moving average
(EMA, decay $0.999$) of the weights, updated every step and used for all
validation, generation, and checkpointing. Weight averaging is a recognized
horizon-free substitute for a decay phase~\cite{anytime,polyak1992}, so the
flat-LR-plus-EMA combination yields checkpoints that are both mutually
comparable and individually well-behaved. \looseness=-1

\paragraph{Objective and masking}
Training uses a completion-only loss mask~\cite{completionmasking}. Each example is
the concatenation of an instruction and its code response, and the cross-entropy
is computed only over the response tokens and the terminating end-of-sequence
token, with all instruction (prompt) tokens and padding positions excluded from
the loss. For the recursive arm, this per-token cross-entropy is averaged over the
$n_{\text{sup}}{=}3$ deep-supervision steps defined in
Section~\ref{subsec:models}; for the two vanilla arms, it is the ordinary
next-token loss. The arms are thus matched on data and masking, but \emph{not} on
the objective (Section~\ref{sec:scope}). \looseness=-1

\paragraph{Training budget and snapshots}
Every arm is trained for a fixed budget of $N{=}40$ epochs over the training split.
The budget was set from a pilot run of the deepest arm (vanilla-20L, the
earliest-plateauing configuration), whose validation loss reached its minimum well
within $40$ epochs and rose thereafter. $40$ epochs thus spans the full arc from
rapid early improvement through plateau to the onset of overfitting for that
arm.\footnote{In the pilot, vanilla-20L reached its validation optimum near epoch
$21$ and a validation-plateau early-stopping criterion fired at epoch $41$; the
main campaign then fixed $N{=}40$ and disabled early stopping so that the entire
trajectory is observed for all arms.} We snapshot each run at nine epochs,
$\{1,2,4,8,12,16,20,28,40\}$: log-spaced early, where model behaviour changes fastest, and more widely spaced late, following the checkpoint-schedule design of the Pythia suite~\cite{pythia}.

\paragraph{Checkpoint selection and reporting}
We select each arm's reported
checkpoint \emph{post hoc} as its best-validation epoch (lowest validation loss).
Training the full budget and selecting post hoc avoids the under-optimization that
patience-based early stopping can induce. For TRM-AR and vanilla-2L, this
best-validation optimum coincides with the budget boundary (validation loss still
decreasing at epoch $40$); we treat their optima as right-censored
(Section~\ref{sec:scope}). Point estimates are reported at each arm's
best-validation checkpoint, and the full nine-snapshot trajectory is reported as
the crossover curve.

\paragraph{Seeds and data regime}
Each arm is trained with three random seeds ($42$, $65$, $108$), controlling
parameter initialization, data ordering, and generation sampling. Because training
performs multiple passes over a fixed corpus rather than a single pass over fresh
data, the study sits in the data-constrained, repeated-data
regime~\cite{muennighoff2023scaling} (Section~\ref{sec:scope}).

\subsection{Evaluation}
\label{subsec:eval}

We score every checkpoint identically, across all arms and snapshots, on two
complementary metric families: teacher-forced fit and free-running generation
(addressing RQ1, RQ2, and RQ4). \looseness=-1

\paragraph{Teacher-forced fit}
We report test-set perplexity, computed from the mean next-token cross-entropy on
the held-out test split under the same completion-only masking used during
training (Section~\ref{subsec:training}). Perplexity is reported as $\exp(\min(\ell, 20))$, where $\ell$ is the mean
completion loss. The clamp guards against numerical outliers at very early
checkpoints and is not active on the reported checkpoints. Since all three arms
share the same tokenizer, vocabulary, test split, and masking, token-level
perplexity is comparable across arms~\cite{paloma}. Deep supervision, however, is a training-time device: for
a fair comparison with the single-pass baselines, we report TRM-AR's perplexity
from its \emph{deployed} forward pass (a single deep recursion, identical to the
computation used to generate) rather than from the mean over the three supervision
steps used during training. This places the reported fit and the free-running
generation metrics on the same inference path.

\paragraph{Free-running generation}
For the free-running measures, each model generates completions for a fixed set of
held-out test prompts, from which we compute two surface~proxies:
\begin{itemize}
  \item \textbf{AST-parse rate}: the fraction of generations whose extracted code
      parses as valid Python under the standard-library \texttt{ast} module. Code
      is extracted from a fenced \texttt{python} block if present, then from a
      bare code fence, and finally from the raw generated text.
  \item \textbf{BLEU-4}: sentence-level BLEU~\cite{bleu} between each generation and
        its reference response, macro-averaged over generations, with whitespace
        tokenization, uniform $n$-gram weights up to $n{=}4$, a brevity penalty, and
        no smoothing. BLEU is a surface-overlap measure known to correlate only
        weakly with functional code correctness~\cite{codebleu}; we therefore report
        BLEU-4 for completeness, but do not use it to rank arms in either contrast.
\end{itemize}
Because the dataset provides no associated unit tests, generations
cannot be executed against a specification, and we therefore report no
execution-based measure of functional correctness such as pass@$k$~\cite{codex}.
We discuss the implications of relying on proxies in Section~\ref{sec:scope}.

\paragraph{LLM-as-judge}
As a generation-quality signal complementary to the surface proxies, we
additionally score completions with an LLM judge~\cite{llmjudge} (Anthropic's Claude
Haiku~4.5). Each (prompt, completion) pair is rated on three integer axes, each on a 0--10
scale, under a fixed rubric: syntax (how syntactically valid the code is, whether
it parses with proper indentation and balanced brackets), logic (how sound its
control flow and variable use are, and whether it is free of obvious bugs or
triviality), and relevance (how directly it implements what the prompt asks). 
The judge is queried at temperature $0$, and completions whose
response cannot be parsed into three scores are discarded; this affected $3$ of
the $81$ evaluated snapshots, one completion each ($0.015\%$ of all judged
completions). Because the judge is shown only the prompt and the completion, it
provides a \emph{reference-free} rating. Moreover, it is applied with the same
model, prompt, and temperature to every arm, so any systematic judgment bias is
shared across arms, and between-arm comparisons are less sensitive to it. The
validity and known biases of this signal are discussed in
Section~\ref{sec:scope}.

\paragraph{Generation configuration}
\label{subsec:gen-config}
For every checkpoint, we sample $5$ completions for each of $50$ held-out test
prompts drawn from the test split with the run's seed ($250$ generations per
snapshot), decoding at temperature $0.8$ with top-$k$ $40$ and up to $256$ new
tokens. Generation reads the raw prompt with no separator, matching the training
format, and each completion is truncated at the first end-of-sequence token.

\subsection{Reproducibility}
\label{subsec:repro}

We report the environment, data, and configuration detail required to reproduce the study.

\paragraph{Reproducibility pins}
First, we record a complete environment freeze. The full set of $129$
installed packages is version-pinned, on a stack of Python~3.10.14 and
PyTorch~2.6.0 (CUDA~12.1), each run on a single NVIDIA A100 80\,GB GPU. Second, we pin the constructed corpus by a
SHA-256 content fingerprint computed over the training, validation, and test
partitions together with the build manifest. Because the corpus derives from a
synthetic source that records no upstream revision identifier, this content
fingerprint, rather than a source-snapshot tag, is what makes the exact splits
recoverable. Third, the build
manifest records the split seed, ratio, resulting counts (Table~\ref{tab:splits}),
and the local GPT-2 tokenizer, so that preprocessing is fully specified. 

\paragraph{Determinism}
Random seeds control parameter initialization, data ordering, and sampling
(Section~\ref{subsec:training}). Because a subset of CUDA operations are
nondeterministic, we make no claim of bit-for-bit reproducibility of individual
runs. Instead we train and report three seeds per arm and read effects through
their across-seed consistency.

\subsection{Compute}
\label{subsec:compute}

The three arms are matched on parameters or on
effective depth but \emph{not} on compute; this
per-step cost addresses RQ3 (Table~\ref{tab:arms}). TRM-AR attains its effective depth of $20$ at essentially the vanilla-2L parameter
budget, but relocates the cost of depth into computation. The
shared block is applied sequentially many times per forward pass, so per-token
compute scales with the depth actually executed rather than with the parameter
count~\cite{parallelloop}. A single TRM-AR forward pass therefore performs
on the order of a $20$-layer computation, and training adds a further factor
because the objective applies deep supervision over $n_{\text{sup}}=3$ steps
(Section~\ref{subsec:training}).

Because this per-step cost is a consequence of the architecture rather than a
controlled variable, we report it as a measured quantity instead of matching it,
and we report several cost indicators rather than a single efficiency
number~\cite{efficiencymisnomer}. We report two: parameters and wall-clock time.
Parameter counts are given in Table~\ref{tab:arms}. For time, floating-point
counts based on parameters understate the sequential cost of recursion, so we
report wall-clock on fixed hardware~\cite{trainlargecompress,schwartz2020green}.
Measured on a single NVIDIA A100 80\,GB GPU, the median time per optimizer step is
approximately $3.0$\,s for TRM-AR, $1.25$\,s for vanilla-20L, and $0.017$\,s for
vanilla-2L. Training one seed to $40$ epochs takes approximately $31$, $14$, and
$4$ hours, respectively. These totals include the evaluation, free-running
generation, and LLM-judging performed at all nine snapshots. For the fast
iso-parameter arm, whose optimizer time alone is under $0.2$ hours, the total is
therefore dominated by fixed per-snapshot overhead, so the total-time ratio is not
a training-only cost ratio. TRM-AR is thus $\approx\!175\times$ slower
per step than its iso-parameter twin and about $2.4\times$
slower than the iso-depth model, despite using $0.6\times$ the latter's
parameters.
We make no claim of a compute- or FLOP-matched comparison.

\section{Results}
\label{sec:results}

\begin{figure*}[t]
\centering
\includegraphics[width=0.8\textwidth]{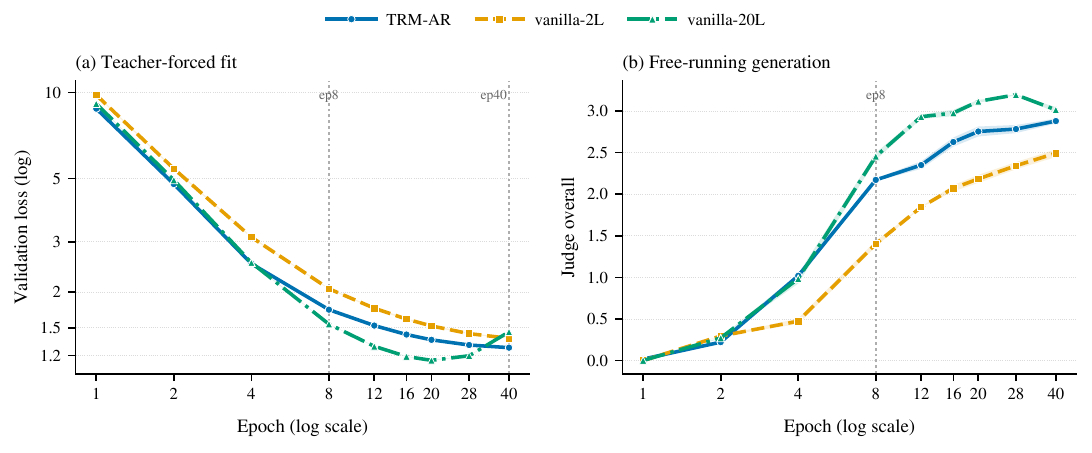}
\caption{Which arm wins depends on when the comparison is read: fit and generation
can rank the three arms differently at a given epoch, and the fit ranking reverses
twice across training. Trajectories of the three arms across the nine snapshot epochs:
(a)~teacher-forced fit (validation loss) and (b)~free-running generation (judge score,
$0$--$10$). Solid lines are the seed-mean; shaded bands are $\pm1$ standard deviation
across the three seeds; the epoch axis is log-scaled. Dotted vertical guides mark the
two fit crossovers in~(a) (epochs~$8$ and~$40$) and, in~(b), the epoch from which
vanilla-20L leads generation ($8$).}
\label{fig:crossover}
\end{figure*}

\begin{figure}[t]
\centering
\includegraphics[width=0.4\textwidth]{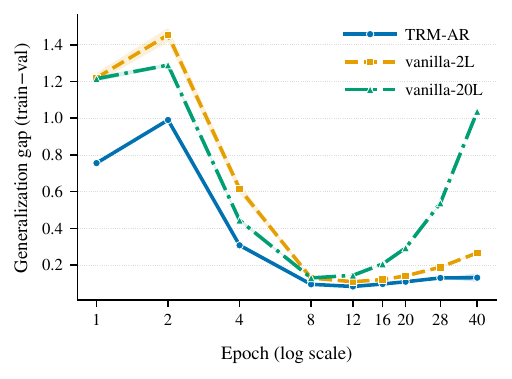}
\caption{After an early transient common to all arms, TRM-AR holds a small, flat
generalization gap, while the iso-depth control overfits after epoch~$20$.
Generalization gap (train $-$ validation) across training; solid lines are the
seed-mean and shaded bands are $\pm1$ standard deviation across the three seeds. The
epoch axis is log-scaled.}
\label{fig:gengap}
\end{figure}

We report point estimates at each arm's \emph{validation-selected} checkpoint (lowest
validation loss, chosen per seed; values are seed means, Table~\ref{tab:bestval}), and
the full nine-snapshot trajectory as a crossover curve
(Figs.~\ref{fig:crossover}--\ref{fig:gengap}). Selecting on validation rather than test
avoids the optimistic bias of test-set selection, so we never report a best-test
number. Because the flat schedule leaves validation loss for TRM-AR and vanilla-2L still
falling at the budget boundary, both arms are selected at epoch~$40$ (except one
TRM-AR seed, selected at epoch~$28$), so their optima are right-censored
(Section~\ref{sec:scope}).
Throughout, we report the direction and across-seed sign-consistency of each effect
rather than significance tests (Section~\ref{sec:scope}).
Results are organized by comparison:
the iso-parameter contrast addresses parameter efficiency and generation quality
(RQ2, RQ4); the iso-depth contrast, the generalization analysis, and the closing
single-snapshot analysis address the across-training question (RQ1); and the
per-step compute cost (RQ3) is measured in Section~\ref{subsec:compute}. Section~\ref{sec:discussion} answers
each question explicitly.
The trajectory itself is the
first result: the fit ranking between TRM-AR and vanilla-20L reverses twice within the
budget (epochs~$8$ and~$40$), so which arm ``fits better'' depends on the epoch at
which the question is asked. We first establish the validation-selected ordering, then show where a single snapshot would misreport the comparison and why.

\begin{table*}[t]
\centering
\renewcommand{\arraystretch}{1.3}
\setlength{\tabcolsep}{10pt}
\begin{threeparttable}
\caption{Validation-Selected Point Estimates for the Three Arms}
\label{tab:bestval}
\begin{tabular}{lccccc}
\toprule
Arm & Sel.\ epoch\tnote{a} & Val.\ loss~$\downarrow$ & Test ppl.~$\downarrow$ &
Judge~$\uparrow$\tnote{b} & Gap$_{\text{tv}}$~$\downarrow$\tnote{c} \\
\midrule
TRM-AR      & $40$ & $1.277_{\pm0.015}$ & $3.53_{\pm0.05}$ & $2.85_{\pm0.06}$ & $\mathbf{0.14}_{\pm0.01\phantom{3}}$ \\
vanilla-2L  & $40$ & $1.379_{\pm0.006}$ & $3.92_{\pm0.02}$ & $2.49_{\pm0.06}$ & $0.27_{\pm0.01\phantom{3}}$ \\
vanilla-20L & $20$ & $\mathbf{1.154}_{\pm0.003}$ & $\mathbf{3.14}_{\pm0.01}$ & $\mathbf{3.12}_{\pm0.02}$ & $0.29_{\pm0.003}$ \\
\bottomrule
\end{tabular}
\begin{tablenotes}[flush]
\footnotesize
\item Seed-mean over $n{=}3$ seeds; subscripts are $\pm1$ s.d.\ across seeds. Arrows give the preferred direction; best value per column in bold. vanilla-2L is the iso-parameter control, vanilla-20L the iso-depth control (Table~\ref{tab:arms}).
\item[a] Best-validation epoch; one TRM-AR seed is selected at epoch~$28$.
\item[b] LLM-judge overall score, $0$--$10$.
\item[c] Train--validation (generalization) gap at the selected epoch.
\end{tablenotes}
\end{threeparttable}
\end{table*}

At their validation-selected checkpoints the ordering is identical on fit and on
judged generation: vanilla-20L leads both, then TRM-AR, then vanilla-2L. At the vanilla-2L parameter budget, TRM-AR closes
about $45\%$ of the gap between vanilla-2L and vanilla-20L on validation loss, and
about $57\%$ on the judge score.

\paragraph{Iso-parameter comparison (RQ2, RQ4)}
At equal parameters, recursion is the stronger arm on every axis except surface
overlap. On fit, TRM-AR has lower validation loss and lower test perplexity at all
nine snapshot epochs, and the sign of the difference is consistent across all three
seeds at every epoch ($9/9$).\footnote{Throughout, $X/9$ denotes the number of the
nine snapshot epochs at which the sign of the between-arm difference is consistent
across all three seeds.} 
The advantage is largest early and shrinks
monotonically without ever reversing: the validation-loss gap narrows from roughly
$1.0$ at epoch~$1$ to about $0.1$ at epoch~$40$. TRM-AR thus reaches any given fit
level earlier, and vanilla-2L narrows the gap but never overtakes TRM-AR. On judged generation, the pattern is opposite in shape but the same in direction. At
epochs~$1$--$2$, a degenerate transient of near-zero-quality output, vanilla-2L
records a marginally higher judge score; from epoch~$4$ onward TRM-AR leads; the between-arm sign is seed-consistent at all nine epochs ($9/9$). The syntax and relevance sub-axes track the overall score, each seed-consistent at
all nine epochs ($9/9$), while the logic sub-axis points the same way but is noisier
($7/9$, seed-consistent at every epoch from~$4$). AST-parse rate favours
TRM-AR from epoch~$2$, though less consistently ($7/9$, and not seed-consistent at
epoch~$40$). BLEU-4 shows no consistent direction across epochs; because BLEU is a weak
surface-overlap measure for code, we read this as a null rather than as evidence
against the judged-quality result. Since the judged-generation advantage begins in
early training rather than at initialization, we scope the iso-parameter finding as
holding \emph{from early training (epoch~$4$) onward}, not throughout.

\paragraph{Iso-depth comparison (RQ1)}
In contrast, at equal effective depth the comparison favours vanilla-20L, and the
trajectory shows where a single snapshot would mislead. On fit there is a
\emph{double crossover} (Fig.~\ref{fig:crossover}a): TRM-AR is ahead very early,
vanilla-20L takes the lead at epoch~$8$ and holds it through epoch~$28$, and the
ordering flips back to TRM-AR at epoch~$40$. The epoch-$40$ reversal is not a
recursion win. It occurs only because vanilla-20L is overfitting late (see
\emph{Generalization and overfitting} below), so its validation loss has risen past that of TRM-AR; under validation selection, vanilla-20L is reported at its epoch-$20$
optimum, where it has the best fit of any arm. On generation, vanilla-20L is
consistently better: it leads the overall judge score from epoch~$8$ onward,
seed-consistent at every epoch from~$8$ ($8/9$ overall). AST-parse rate, by
contrast, does not reliably separate the two arms (neither leads with a
seed-consistent sign across epochs), so the generation gap in this contrast rests on
the judge rather than on surface validity. The iso-depth contrast therefore shows that weight-sharing yields no fit or generation advantage over a model with $\approx\!10\times$ the
non-embedding parameters: at its validation-selected optimum the iso-depth control
generates and fits better, and weight-sharing does not close that gap at this scale. \looseness=-1

\paragraph{Generalization and overfitting (RQ1, RQ2)}
The three arms separate sharply on the train--validation (generalization) gap,
ordered TRM-AR, then vanilla-2L, then vanilla-20L (Fig.~\ref{fig:gengap}). At
epoch~$40$, the gap is $0.13$ for TRM-AR ($0.14$ at its validation-selected
checkpoints in Table~\ref{tab:bestval}, where one seed is read at epoch~$28$),
$0.27$ for vanilla-2L, and $1.04$ for vanilla-20L. TRM-AR carries the highest
training loss of the three yet the lowest validation loss of the three arms at that
epoch: it fits the training data least and generalizes best, a pattern consistent
with regularization that we return to in Section~\ref{sec:discussion}. vanilla-20L
shows pronounced overfitting: its validation loss reaches its minimum at epoch~$20$
($1.15$) and rises to epoch~$40$ ($1.45$) while its training loss keeps falling, so
its gap widens sharply (to $1.04$). vanilla-2L is intermediate: its gap widens from
epoch~$20$ to~$40$ ($0.14\rightarrow0.27$) even as its validation loss continues to
fall. This is the onset of overfitting, but without the validation-loss reversal that vanilla-20L undergoes. The cross-seed spread is tight (e.g.\ the three vanilla-20L
epoch-$40$ validation losses fall within $\pm0.002$), so the ordering is a robust
property of the architectures, not a single-seed artifact. Within this budget,
TRM-AR thus resists the overfitting that vanilla-20L suffers, and it generalizes
more stably even than vanilla-2L, its iso-parameter control.

\paragraph{Fit--generation decoupling (RQ1, RQ4)}
The clearest instance of fit diverging from generation is \emph{within} vanilla-20L. From epoch~$20$ to epoch~$40$ its teacher-forced fit degrades
markedly (validation loss $1.15\rightarrow1.45$; test perplexity
$3.14\rightarrow4.22$) while its free-running generation declines only slightly: the
overall judge score falls from $3.12$ to $3.01$, a small but seed-consistent decline
(each of the three seeds declines by less than $0.13$ on the $0$--$10$ scale).
Training past its optimum therefore harms what the model predicts under teacher
forcing far more than what it generates: fit and generation decouple within a single
model and a single training run.

\paragraph{Single-snapshot artifacts (RQ1)}
This within-model decoupling, together with the iso-depth double crossover above, shows directly how single-checkpoint readings of these runs mislead.
A mid-training snapshot
(epochs~$8$--$28$) would report ``the larger model fits better.'' The epoch-$40$
snapshot would report ``the recursive model fits better.'' Neither single reading is
a faithful summary, and the second is an overfitting artifact: TRM-AR regains the
fit lead at epoch~$40$ only because vanilla-20L has degraded past its optimum,
not because recursion has improved relative to it. Validation-based checkpoint
selection together with the full trajectory resolves the contradictory snapshots
into a consistent result, in which each arm is read at its
validation-selected optimum and the late crossing is recognized as degradation
rather than a win.

\section{Discussion}
\label{sec:discussion}

Taken together, the results give a two-sided answer: which model wins is time-indexed, and validation selection across the full trajectory resolves it into a stable comparison. We first summarize how the results answer each research question, then turn to interpretation and implications.

\textbf{RQ1:} Read across the budget, the comparison against the iso-depth control
names no single winner. The fit ranking between TRM-AR and vanilla-20L reverses twice
(Section~\ref{sec:results}), and the late reversal is an overfitting artifact rather
than a recursion win. Validation selection resolves the contradiction, reporting
vanilla-20L at its epoch-$20$ optimum, where it both fits and generates better than
either other arm. The recursive configuration's remaining advantage over the iso-depth control is therefore resistance to
overfitting, not greater capability.
\textbf{RQ2:} At equal parameters, TRM-AR attains lower validation loss and test
perplexity than vanilla-2L at every snapshot epoch, and it generalizes better,
carrying a smaller train--validation gap. These gains are a
property of the TRM-AR configuration, not of recursion in isolation.
\textbf{RQ3:} The efficiency is in parameters, not in compute. Realizing an effective
depth of $20$ by looping a two-layer block costs roughly $175\times$ the per-step time
of the iso-parameter control and about $2.4\times$ that of the iso-depth control
(Section~\ref{subsec:compute}); recursion relocates the cost of depth into sequential
computation.
\textbf{RQ4:} At equal parameters recursion also improves \emph{free-running}
generation, as scored by the LLM judge, from epoch~$4$ onward and consistently across
seeds. BLEU-4, a weak surface measure for code, shows no stable difference
(Section~\ref{sec:results}).

\paragraph{Decoupling and regularization}
Prior work distinguishes teacher-forced fit from free-running generation as separate
regimes, in which strength in one does not entail strength in the
other~\cite{nexttokenpitfalls,exposurebias}. Our results give a concrete instance
within vanilla-20L, whose fit degrades markedly as it trains past its epoch-$20$
optimum while its judged generation declines only slightly
(Section~\ref{sec:results}). We describe this as a \emph{decoupling} and deliberately
do not attribute it to a specific named mechanism such as exposure bias. The claim is
descriptive, namely that a single model's two competences degrade at sharply different
rates. A second pattern runs through the same contrasts: parameter sharing behaves
like an implicit regularizer. TRM-AR carries
the \emph{highest} training loss of the three arms yet the \emph{smallest}
generalization gap (Section~\ref{sec:results}), consistent with the view that
cross-layer parameter sharing regularizes training and improves
generalization~\cite{albert}. We frame this as an observation rather than a controlled
result, because the effect is entangled with depth, deep supervision, and
architecture. The overfitting of the iso-depth control is in turn consistent with the
repeated-data regime, in which repeated tokens and excess parameters both lose
value~\cite{muennighoff2023scaling}.

\paragraph{Relation to prior work}
The two contrasts land differently against the literature. The iso-parameter result
strengthens the recursive-efficiency line by carrying it from reasoning accuracy at
convergence to free-running code generation tracked across training. The equal-parameter gain is also consistent in size with the closest prior comparison: Saunshi et
al.~\cite{saunshi} report looped models covering roughly a third to a half of the
perplexity gap between their parameter- and depth-matched controls, and TRM-AR closes
about $45\%$ of the corresponding validation-loss gap. Concurrent work adapting TRM to
autoregressive decoding reports no reliable gains on synthetic character-level
algorithmic tasks under compute-matched comparisons~\cite{artrm}. That null does not
conflict with our equal-parameter result, because the two studies match different
quantities (parameters and effective depth versus compute), evaluate different tasks,
and train different configurations, ours retaining TRM's deep-supervision recipe. We
do not read the iso-depth result as contradicting the field's efficiency claims, which
concern different scales, tasks, and metrics. At $\sim\!28$M on code,  the recursive configuration's
benefit relative to the iso-depth control is resistance to overfitting under a limited
data budget, not raw capability.

\paragraph{Implications for practice and evaluation}
For a practitioner building a small code model under a fixed parameter budget and
limited data, the results favour recursion with deep supervision as the default
choice. It fits and generates better than an equal-parameter plain transformer and
resists overfitting, provided the roughly $175\times$ per-step compute cost is
acceptable for the deployment.
Where inference latency
dominates and parameters are cheap, a larger plain transformer, selected at its
validation optimum, remains the stronger choice. 
The results also carry a methodological point. A single-checkpoint fit comparison
between TRM-AR and vanilla-20L names a different winner at different epochs; reading
each arm at its validation-selected optimum, against the full trajectory, yields one
stable ranking. This pattern echoes the rationale for released checkpoint suites in
training-dynamics research~\cite{pythia}: the point during training at which a
comparison is made can change its conclusion, so recursive models should be evaluated
on fit and generation jointly, across training. 

\section{Limitations}
\label{sec:scope}

This study is a controlled comparison of three architectures on a shared
instruction-to-code corpus across training, not a benchmark of absolute
code-generation ability. Its limitations follow the standard validity dimensions.

\textbf{Internal validity.} TRM-AR differs from its iso-parameter control on two axes
at once: it realizes depth by recursion \emph{and} it is trained with deep supervision,
whereas vanilla-2L uses ordinary next-token loss. We do not ablate the two, and an
independent analysis of HRM, the model TRM simplifies, identified deep supervision
rather than recursion as the apparent primary driver of that model's
gains~\cite{arcprize_hrm}. The equal-parameter findings therefore credit the TRM-AR
configuration as a whole, which bounds the mechanistic claim but not the efficiency
finding. \looseness=-1

\textbf{Construct validity.} The corpus has no unit tests and pass@$k$ is uninformative
at this scale, so we report no execution-based correctness~\cite{codex} and rely on
proxies: AST-parse rate, BLEU-4, and an LLM judge. These are imperfect: BLEU correlates
only weakly with functional correctness~\cite{codebleu}, and the judge is uncalibrated
and unvalidated against execution at this scale~\cite{judgeexecagreement}. We therefore
read the judge comparatively, as an ordering between arms at matched training points,
and treat convergent movement across the proxies as signal.

\textbf{External validity.} The corpus is synthetic, templated, and LLM-generated
(Section~\ref{subsec:dataset}). This regularity makes a controlled comparison learnable
but differs from human-authored code, so the absolute levels of our proxies and their
transfer to naturally occurring code are not established here. Training also performs
many passes over a fixed corpus, so conclusions about across-training behaviour are
specific to this data-constrained, repeated-data regime~\cite{muennighoff2023scaling}.

\textbf{Conclusion validity.} Each arm is trained with three seeds, too small a sample
for statistical significance, so we report the direction, raw magnitude, and
sign-consistency of each effect rather than $p$-values or standardized effect sizes.
The fixed $40$-epoch budget also right-censors the two parameter-lean arms, whose
validation loss is still decreasing at epoch~$40$ (Section~\ref{subsec:training}); 
we therefore claim the iso-depth ordering only \emph{within this budget}. \looseness=-1

Within these constraints, the study provides a controlled,
seed-replicated characterization of teacher-forced fit and free-running generation
across training.

\section{Conclusion}
\label{sec:conclusion}

We compared an autoregressive Tiny Recursive Model against parameter- and
depth-matched transformer baselines on natural-language-to-Python code generation,
tracking teacher-forced fit and free-running generation across training and three seeds
rather than at a single converged checkpoint. Read at
different checkpoints, the comparison points to different winners; selecting each arm at its validation optimum and reading the full
trajectory resolves this into a two-sided result. At equal parameters, recursion is the
more parameter-efficient design, fitting, generating, and generalizing better across
seeds. At equal effective depth, a larger plain transformer fits and generates better at
its optimum. The recursive configuration retains only resistance to overfitting, not greater
capability, and its efficiency is in parameters, rather than compute.

In this controlled study of small recursive code generation models, architecture rankings changed substantially across training, indicating that trajectory-aware evaluation can provide a more reliable assessment than single-checkpoint comparisons in similar settings. We therefore advocate trajectory-based evaluation for recursive models, and suggest it may be valuable more broadly for emerging small-scale architectures whose learning dynamics may differ substantially.

Several directions remain for future work. First, a deep-supervised shallow transformer would help disentangle the effects of recursive parameter sharing from those of deep supervision. Second, extending the study to larger models, more diverse code-generation benchmarks, and execution-based metrics such as pass@k would establish the generality of the observed crossover behaviour. Finally, evaluating additional recursive and weight-sharing architectures would determine whether trajectory-dependent rankings are a broader property of recursive models or specific to the TRM studied here. \looseness=-1

\ifanon\else
\fi

\bibliographystyle{IEEEtran}
\bibliography{references}

\begin{thebibliography}{10}
\providecommand{\url}[1]{#1}
\csname url@samestyle\endcsname
\providecommand{\newblock}{\relax}
\providecommand{\bibinfo}[2]{#2}
\providecommand{\BIBentrySTDinterwordspacing}{\spaceskip=0pt\relax}
\providecommand{\BIBentryALTinterwordstretchfactor}{4}
\providecommand{\BIBentryALTinterwordspacing}{\spaceskip=\fontdimen2\font plus
\BIBentryALTinterwordstretchfactor\fontdimen3\font minus \fontdimen4\font\relax}
\providecommand{\BIBforeignlanguage}[2]{{%
\expandafter\ifx\csname l@#1\endcsname\relax
\typeout{** WARNING: IEEEtran.bst: No hyphenation pattern has been}%
\typeout{** loaded for the language `#1'. Using the pattern for}%
\typeout{** the default language instead.}%
\else
\language=\csname l@#1\endcsname
\fi
#2}}
\providecommand{\BIBdecl}{\relax}
\BIBdecl

\bibitem{codegensurvey}
J.~Jiang, F.~Wang, J.~Shen, S.~Kim, and S.~Kim, ``A survey on large language models for code generation,'' \emph{ACM Transactions on Software Engineering and Methodology (TOSEM)}, 2025.

\bibitem{vaswani}
A.~Vaswani, N.~Shazeer, N.~Parmar, J.~Uszkoreit, L.~Jones, A.~N. Gomez, L.~Kaiser, and I.~Polosukhin, ``Attention is all you need,'' in \emph{Advances in Neural Information Processing Systems (NeurIPS)}, 2017.

\bibitem{chinchilla}
J.~Hoffmann, S.~Borgeaud, A.~Mensch \emph{et~al.}, ``Training compute-optimal large language models,'' in \emph{Proc. Advances in Neural Information Processing Systems (NeurIPS)}, 2022.

\bibitem{codegen}
E.~Nijkamp, B.~Pang, H.~Hayashi, L.~Tu, H.~Wang, Y.~Zhou, S.~Savarese, and C.~Xiong, ``{CodeGen}: An open large language model for code with multi-turn program synthesis,'' in \emph{Proc. International Conference on Learning Representations (ICLR)}, 2023.

\bibitem{codet5plus}
Y.~Wang, H.~Le, A.~D. Gotmare, N.~D.~Q. Bui, J.~Li, and S.~C.~H. Hoi, ``{CodeT5+}: Open code large language models for code understanding and generation,'' in \emph{Proc. Conference on Empirical Methods in Natural Language Processing (EMNLP)}, 2023, pp. 1069--1088.

\bibitem{alphacode}
Y.~Li, D.~Choi, J.~Chung, N.~Kushman \emph{et~al.}, ``Competition-level code generation with {AlphaCode},'' \emph{Science}, vol. 378, no. 6624, pp. 1092--1097, 2022.

\bibitem{starcoder}
R.~Li, L.~B. Allal, Y.~Zi \emph{et~al.}, ``{StarCoder}: May the source be with you!'' \emph{Transactions on Machine Learning Research (TMLR)}, 2023.

\bibitem{schwartz2020green}
R.~Schwartz, J.~Dodge, N.~A. Smith, and O.~Etzioni, ``Green {AI},'' \emph{Communications of the ACM}, vol.~63, no.~12, pp. 54--63, 2020, arXiv:1907.10597.

\bibitem{muennighoff2023scaling}
N.~Muennighoff, A.~M. Rush, B.~Barak, T.~Le~Scao, A.~Piktus, N.~Tazi, S.~Pyysalo, T.~Wolf, and C.~Raffel, ``Scaling data-constrained language models,'' in \emph{Advances in Neural Information Processing Systems (NeurIPS)}, 2023, arXiv:2305.16264.

\bibitem{ut}
M.~Dehghani, S.~Gouws, O.~Vinyals, J.~Uszkoreit, and L.~Kaiser, ``Universal transformers,'' in \emph{ICLR}, 2019, arXiv:1807.03819.

\bibitem{albert}
Z.~Lan \emph{et~al.}, ``{ALBERT}: A lite {BERT} for self-supervised learning of language representations,'' in \emph{Proc. International Conference on Learning Representations (ICLR)}, 2020.

\bibitem{trm}
A.~Jolicoeur-Martineau, ``Less is more: Recursive reasoning with tiny networks,'' 2025, arXiv:2510.04871.

\bibitem{hrm}
G.~Wang, J.~Li, Y.~Sun \emph{et~al.}, ``Hierarchical reasoning model,'' 2025, arXiv:2506.21734.

\bibitem{teacherforcing}
R.~J. Williams and D.~Zipser, ``A learning algorithm for continually running fully recurrent neural networks,'' \emph{Neural Computation}, vol.~1, no.~2, pp. 270--280, 1989.

\bibitem{saunshi}
N.~Saunshi, N.~Dikkala, Z.~Li, S.~Kumar, and S.~J. Reddi, ``Reasoning with latent thoughts: On the power of looped transformers,'' in \emph{International Conference on Learning Representations (ICLR)}, 2025, arXiv:2502.17416.

\bibitem{professorforcing}
A.~Lamb, A.~Goyal, Y.~Zhang, S.~Zhang, A.~Courville, and Y.~Bengio, ``Professor forcing: A new algorithm for training recurrent networks,'' in \emph{Advances in Neural Information Processing Systems (NeurIPS)}, 2016.

\bibitem{nexttokenpitfalls}
G.~Bachmann and V.~Nagarajan, ``The pitfalls of next-token prediction,'' in \emph{ICML}, 2024, arXiv:2403.06963.

\bibitem{exposurebias}
K.~Arora, L.~E. Asri, H.~Bahuleyan, and J.~C.~K. Cheung, ``Why exposure bias matters: An imitation learning perspective of error accumulation in language generation,'' in \emph{Findings of the Association for Computational Linguistics: ACL 2022}, 2022, pp. 700--710, arXiv:2204.01171.

\bibitem{pythia}
S.~Biderman, H.~Schoelkopf, Q.~Anthony \emph{et~al.}, ``Pythia: A suite for analyzing large language models across training and scaling,'' in \emph{ICML}, 2023, arXiv:2304.01373.

\bibitem{tinycodes}
{Nam Pham}, ``tiny-codes,'' Hugging Face Datasets, 2023, \url{https://huggingface.co/datasets/nampdn-ai/tiny-codes}, accessed 2026-07-13.

\bibitem{herrmann2024rethinking}
M.~Herrmann, F.~J.~D. Lange, K.~Eggensperger, G.~Casalicchio, M.~Wever, M.~Feurer, D.~R{\"u}gamer, E.~H{\"u}llermeier, A.-L. Boulesteix, and B.~Bischl, ``Position: Why we must rethink empirical research in machine learning,'' in \emph{Proceedings of the 41st International Conference on Machine Learning (ICML)}, 2024, arXiv:2405.02200.

\bibitem{llmjudge}
L.~Zheng, W.-L. Chiang, Y.~Sheng, S.~Zhuang, Z.~Wu, Y.~Zhuang, Z.~Lin, Z.~Li, D.~Li, E.~P. Xing, H.~Zhang, J.~E. Gonzalez, and I.~Stoica, ``Judging {LLM-as-a-Judge} with {MT-Bench} and {Chatbot Arena},'' in \emph{Advances in Neural Information Processing Systems (NeurIPS)}, 2023, arXiv:2306.05685.

\bibitem{artifact}
A.~Sirivella, A.~Newaz, and G.~Melo, ``Evaluating tiny recursive models across training for code generation,'' \url{https://doi.org/10.5281/zenodo.21347848}, 2026.

\bibitem{relaxedrecursive}
S.~Bae \emph{et~al.}, ``Relaxed recursive transformers: Effective parameter sharing with layer-wise {LoRA},'' 2024, arXiv:2410.20672.

\bibitem{geiping}
J.~Geiping, S.~McLeish \emph{et~al.}, ``Scaling up test-time compute with latent reasoning: A recurrent depth approach,'' 2025, arXiv:2502.05171.

\bibitem{ouro}
R.-J. Zhu, Z.~Wang, K.~Hua \emph{et~al.}, ``Scaling latent reasoning via looped language models,'' 2025, arXiv:2510.25741.

\bibitem{artrm}
P.~Rauba, C.~Fanconi, and M.~van~der Schaar, ``Tiny autoregressive recursive models,'' \emph{arXiv preprint arXiv:2603.08082}, 2026, iCLR 2026 Workshop on AI with Recursive Self-Improvement.

\bibitem{knowledgecapacity}
Z.~Allen-Zhu and Y.~Li, ``Physics of language models: Part 3.3, knowledge capacity scaling laws,'' 2024, arXiv:2404.05405.

\bibitem{scheduledsampling}
S.~Bengio, O.~Vinyals, N.~Jaitly, and N.~Shazeer, ``Scheduled sampling for sequence prediction with recurrent neural networks,'' in \emph{Advances in Neural Information Processing Systems (NeurIPS)}, 2015, arXiv:1506.03099.

\bibitem{seqleveltraining}
M.~Ranzato, S.~Chopra, M.~Auli, and W.~Zaremba, ``Sequence level training with recurrent neural networks,'' in \emph{International Conference on Learning Representations (ICLR)}, 2016, arXiv:1511.06732.

\bibitem{codex}
M.~Chen, J.~Tworek, H.~Jun \emph{et~al.}, ``Evaluating large language models trained on code,'' 2021, arXiv:2107.03374.

\bibitem{phi1}
S.~Gunasekar, Y.~Zhang, J.~Aneja \emph{et~al.}, ``Textbooks are all you need,'' 2023, arXiv:2306.11644.

\bibitem{gpt2}
A.~Radford \emph{et~al.}, ``Language models are unsupervised multitask learners,'' 2019.

\bibitem{swiglu}
N.~Shazeer, ``Glu variants improve transformer,'' 2020, arXiv:2002.05202.

\bibitem{rmsnorm}
B.~Zhang and R.~Sennrich, ``Root mean square layer normalization,'' in \emph{Proc. Advances in Neural Information Processing Systems (NeurIPS)}, 2019.

\bibitem{rope}
J.~Su, Y.~Lu, S.~Pan, A.~Murtadha, B.~Wen, and Y.~Liu, ``{RoFormer}: Enhanced transformer with rotary position embedding,'' \emph{Neurocomputing}, vol. 568, p. 127063, 2024, arXiv:2104.09864.

\bibitem{kaplan}
J.~Kaplan, S.~McCandlish, T.~Henighan \emph{et~al.}, ``Scaling laws for neural language models,'' 2020, arXiv:2001.08361.

\bibitem{adamw}
I.~Loshchilov and F.~Hutter, ``Decoupled weight decay regularization,'' in \emph{Proc. International Conference on Learning Representations (ICLR)}, 2019.

\bibitem{wsd}
S.~Hu, Y.~Tu, X.~Han \emph{et~al.}, ``Minicpm: Unveiling the potential of small language models with scalable training strategies,'' 2024, arXiv:2404.06395.

\bibitem{anytime}
A.~Meterez, P.~A. Nair, D.~Morwani, C.~Pehlevan, and S.~Kakade, ``Anytime pretraining: Horizon-free learning-rate schedules with weight averaging,'' 2026, arXiv:2602.03702.

\bibitem{polyak1992}
B.~T. Polyak and A.~B. Juditsky, ``Acceleration of stochastic approximation by averaging,'' \emph{SIAM Journal on Control and Optimization}, vol.~30, no.~4, pp. 838--855, 1992.

\bibitem{completionmasking}
Z.~Shi, A.~X. Yang \emph{et~al.}, ``Instruction tuning with loss over instructions,'' in \emph{Proc. Advances in Neural Information Processing Systems (NeurIPS)}, 2024.

\bibitem{paloma}
I.~Magnusson, A.~Bhagia, V.~Hofmann \emph{et~al.}, ``Paloma: A benchmark for evaluating language model fit,'' 2023, arXiv:2312.10523.

\bibitem{bleu}
K.~Papineni \emph{et~al.}, ``{BLEU}: a method for automatic evaluation of machine translation,'' in \emph{Proc. Association for Computational Linguistics (ACL)}, 2002.

\bibitem{codebleu}
S.~Ren, D.~Guo, S.~Lu, L.~Zhou, S.~Liu, D.~Tang, N.~Sundaresan, M.~Zhou, A.~Blanco, and S.~Ma, ``{CodeBLEU}: a method for automatic evaluation of code synthesis,'' 2020, arXiv:2009.10297.

\bibitem{parallelloop}
B.~Wu, M.~Chen, X.~Luo \emph{et~al.}, ``Parallel loop transformer for efficient test-time computation scaling,'' 2025, arXiv:2510.24824.

\bibitem{efficiencymisnomer}
M.~Dehghani, A.~Arnab, L.~Beyer, A.~Vaswani, and Y.~Tay, ``The efficiency misnomer,'' in \emph{ICLR}, 2022, arXiv:2110.12894.

\bibitem{trainlargecompress}
Z.~Li, E.~Wallace, S.~Shen, K.~Lin, K.~Keutzer, D.~Klein, and J.~E. Gonzalez, ``Train large, then compress: Rethinking model size for efficient training and inference of transformers,'' in \emph{ICML}, 2020, arXiv:2002.11794.

\bibitem{arcprize_hrm}
{ARC Prize Foundation}, ``The hidden drivers of {HRM}'s performance on {ARC-AGI},'' \url{https://arcprize.org/blog/hrm-analysis}, 2025, online; accessed 2026-07-04.

\bibitem{judgeexecagreement}
G.~Crupi, R.~Tufano, A.~Velasco, A.~Mastropaolo, D.~Poshyvanyk, and G.~Bavota, ``On the effectiveness of {LLM}-as-a-judge for code generation and summarization,'' \emph{IEEE Transactions on Software Engineering}, 2025, arXiv:2507.16587.

\end{thebibliography}

\end{document}